\documentclass[10pt,twocolumn,letterpaper]{article}

\usepackage{cvpr}
\usepackage{times}
\usepackage{epsfig}
\usepackage{graphicx}
\usepackage{amsmath}
\usepackage{amssymb}
\usepackage{xcolor}
\usepackage{subcaption}
\usepackage{amsthm}
\usepackage{algorithm}
\usepackage{algpseudocode}
\usepackage{booktabs}
\usepackage[numbers]{natbib}
\usepackage{lipsum}
\usepackage{array}

\usepackage[breaklinks=true,bookmarks=false]{hyperref}
\usepackage{cleveref}

\cvprfinalcopy 

\def\cvprPaperID{****} 

\theoremstyle{definition}
\newtheorem{definition}{Definition}[section]

\DeclareMathOperator{\sign}{sign}
\DeclareMathOperator{\avg}{avg}

\DeclareMathOperator*{\mean}{mean}

\begin{document}

\title{Increasing the adversarial robustness and explainability \\ of capsule networks with $\gamma$-capsules}

\author{David Peer\\
University of Innsbruck\\
Austria\\
{\tt\small d.peer@uibk.ac.at}
\and
Sebastian Stabinger\\
University of Innsbruck\\
Austria\\
{\tt\small sebastian@stabinger.name}
\and
Antonio Rodr\'{i}guez-S\'{a}nchez\\
University of Innsbruck\\
Austria\\
{\tt\small antonio.rodriguez-sanchez@uibk.ac.at}
}

\maketitle

\begin{abstract}
   In this paper we introduce a new
   inductive bias for capsule networks and call networks that use
   this prior $\gamma$-capsule networks. Our 
   inductive bias that is inspired by TE neurons of the inferior temporal cortex increases the adversarial robustness and the  explainability of capsule networks. 
   A theoretical framework with formal definitions of $\gamma$-capsule networks
   and metrics for evaluation are also provided. Under our framework we show that
   common capsule networks do not necessarily make use of this inductive bias. For this reason we introduce 
   a novel routing algorithm and use a different training algorithm to be 
   able to implement $\gamma$-capsule networks.
   We then show experimentally that $\gamma$-capsule networks
   are indeed more transparent and more robust against adversarial attacks than 
   regular capsule networks.
\end{abstract}

\section{Introduction}
Animals and humans are born with a highly structured brain that allows 
them to function right after birth, this fact may be due to the presence of an \emph{inductive
bias} \citep{critique-pure-learning} acquired through evolution. 
This inductive
bias together with learning is advantageous over \emph{pure-learning}, because it allows animals 
to learn specific things very quickly. Analogous approaches may also
accelerate and improve the progress in the current state of Artificial Neural Networks (ANNs) \citep{critique-pure-learning}. One very successful example are 
Convolutional Neural Networks (CNNs) \cite{cnn}, which are motivated by 
the receptive field of neurons from the visual cortex as introduced in Fukushima's Neocognitron \citep{neocognitron-self-org-nn}. 
In CNNs, this inductive bias exploits the fact that input images are translational invariant,
largely reducing the number of parameters to be learned and increasing the overall classification performance of the network.
\begin{figure}[t]
   \centering

   \begin{subfigure}{.45\linewidth}
      \centering
      \includegraphics[width=0.5\linewidth]{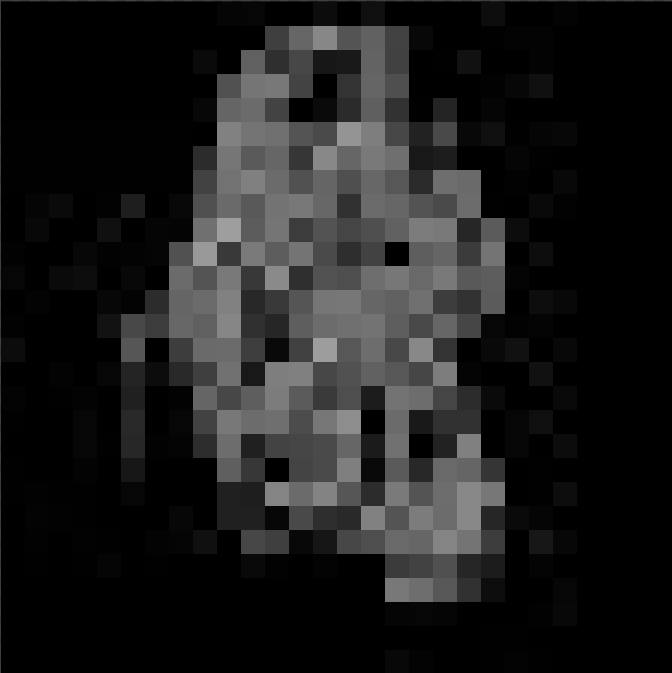}
      \caption{}
      \label{fig:feature_samples_a}
   \end{subfigure}
   \begin{subfigure}{.45\linewidth}
      \centering
      \includegraphics[width=0.5\linewidth]{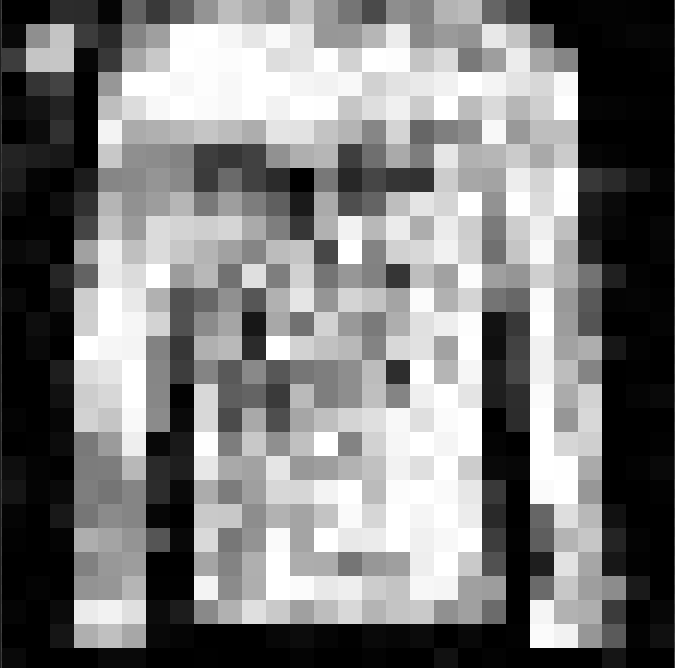}
      \caption{}
      \label{fig:feature_samples_b}
   \end{subfigure}

   \caption{Input features that are generated to activate single $\gamma$-capsules. Input (a) are features that 
   activate $\gamma$-capsule $17$ of the first hidden layer.
   Input (b) are features that activate the output $\gamma$-capsule \emph{pullover}.}
   \label{fig:feature_samples}

\end{figure}
In this paper we introduce a new inductive bias for capsule networks 
that is inspired by the biological visual neurons in area TE of the inferior temporal cortex (IT).
TE neurons encode moderately  
complex and comprehensible object features which are much more complex than just the edges, corners and curvatures analyzed by the neurons from areas V1, V2 and V4 of the visual cortex  \citep{inferotemporal-cortex}.
TE neurons encode 
object parts such that the read out of TE neurons seem to combine information 
of multiple TE neurons to encode an explicit object representation.
Our work is inspired by this hypothesis from research in neurophysiology, which is implemented in the form of a new type of capsule networks which we call \emph{$\gamma$-capsule networks}. 
A \emph{$\gamma$-capsule} represents a \emph{human comprehensible} object or a \emph{human comprehensible} moderately complex part of an object in contradiction to classical capsules which not necessarily encodes human comprehensible objects. During inference, a $\gamma$-capsule is active 
if and only if the feature that it represents exists in the current input. 
$\gamma$-capsules of upper level layers are combinations of lower level $\gamma$-capsules.
An example is shown in \cref{fig:feature_samples} where we can see that a lower level $\gamma$-capsule represents the body of 
the classes \emph{pullover, t-shirt, dress, coat} or \emph{shirt}. To classify the pullover that is shown in \cref{fig:feature_samples_b} correctly, multiple lower level capsules need to be combined.

As the trust in Artificial Intelligence (AI) methods is increasing in critical environments 
such as health care, autonomous cars or finance \& economy, it is important to make sure 
that those models are secure and that they are comprehensible for people, thus, the recent interest in explainable AI. Otherwise, the use of AI may give rise to life-threatening situations as recent work has shown \cite{tesla-autopilot}, where applying minor changes on the road, lead to critical failures of automatic lane recognition systems in autonomous cars. We will show that a $\gamma$-capsule encodes features that are comprehensible for humans.
Those features can be generated visually such that
we can analyze the features that activate units.
Our network structure combines information from lower level 
capsules to produce upper level capsules, solving the problem of assigning parts to wholes \cite{transforming-autoencoders}. 
As shown in \cref{fig:feature_samples}, the connectivity between different layers is created in an explainable way. We will further show that our approach is also very robust against adversarial attacks. \citet{adversarial-examples-are-not-bugs} has shown that adversarial examples can be attributed to features that are highly predictive but incomprehensible for humans (\emph{useful non-robust features}).
As a $\gamma$-capsule encodes only comprehensible features, it will be robust against adversarial attacks.

To be able to implement this inductive bias we introduce a theoretical framework for 
$\gamma$-capsule networks. This framework includes a formal definition 
and metrics to measure the prior that is needed for $\gamma$-capsule networks. 
Using this framework we show that common state of the art capsule networks are not 
$\gamma$-capsule networks. Therefore, we introduce a novel routing algorithm 
called \emph{scaled-distance-agreement} (SDA). We show experimentally that this 
algorithm produces a $\gamma$-capsule network and 
that those networks are more robust against adversarial attacks than CNNs or 
classical capsule networks. We also show that in contrast to classical capsules,  
$\gamma$-capsules are comprehensible for humans. The novel contributions of this paper include: (1) $\gamma$-capsule networks, 
(2) a theoretical framework for $\gamma$-capsules,
(3) SDA-routing to implement $\gamma$-capsules, 
and (4) a novel method to analyze $\gamma$-capsule networks.

The paper is structured as follows: In \cref{sec:related_work} we describe related work. 
A formal definition and metrics for $\gamma$-capsule networks are introduced in 
\cref{sec:framework}. In \cref{sec:implementation} we show how 
a $\gamma$-capsule network can be implemented. In the experimental evaluation \cref{sec:evaluation} we 
compare $\gamma$-capsule networks with the most commonly state of the art capsule networks 
used for supervised learning: matrix capsules with expectation maximization (EM) routing 
and capsule networks with routing-by agreement (RBA). 
We will finish this paper with a discussion on the results and their implications.

\section{Related work}\label{sec:related_work}

\citet{transforming-autoencoders} introduced capsules and the idea that a capsule represents an object or part of an object in a parse tree. In that same work, the authors also
showed how such a capsule can be trained by backpropagating the
difference between the actual and the target outputs. Later,  
\citet{dynamic-routing} and \citet{matrix-capsules} introduced 
routing algorithms to connect capsules of different layers for supervised learning.
Capsule networks have been recently used for different applications such as 
lung cancer screening \cite{lung-cancer-screening},  detecting actions in videos \cite{video-capsnet}
or object classification in 3D point clouds \cite{3d-point-capsules}. \citet{deepcaps} 
created a deep capsule network resulting in state-of-the-art performance on SVHN, CIFAR10
and fashionMNIST. An unsupervised version of capsule networks was trained by \citet{stacked-capsule-autoencoders}.
Previous work on explainable AI has shown that it is not possible to directly sample human comprehensible images to activate a single unit of an ANN. Methods that generate inputs to activate single units in ANNs need to be constrained such that they resemble natural images, otherwise unrealistic inputs are produced \cite{deep-nn-easily-fooled, deep-inside}. 
In order to avoid this problem,  \citet{object-detectors-emerge} start from correctly classified images and simplify this image such that it keeps as little information 
as possible but still produces a large classification score.
Recent research on adversarial attacks shows that those attacks
exploit \emph{useful non-robust features}, because they are highly 
predictive but incomprehensible for humans \cite{adversarial-examples-are-not-bugs}. The authors of 
this work proved this claim experimentally using a robust CNN that was trained with the method  introduced by \citet{resistant-to-adversraial-attacks}.  In the case of capsule networks,  \citet{capsule-adversarial-attacks} has already shown that they can be fooled by adversarial attacks as easily as CNNs. In order to overcome this limitation, \citet{detecting-adversaries} used the reconstruction network of capsule networks to detect adversarial examples. Unfortunately, this novel method can still be fooled with more advanced attacks such as reconstructive attacks  \citep{detecting-adversaries}.

\section{A framework for $\gamma$-capsule networks}\label{sec:framework}

In this section we provide a formal definition of  $\gamma$-capsule networks  
and present the metrics that measure whether a capsule network 
is also a $\gamma$-capsule network. In order to achieve this, we will adapt the $\rho$-useful 
features and the \emph{$\gamma$-robustly useful features} presented by 
\citet{adversarial-examples-are-not-bugs} to a multi-class setting.

\subsection{Definitions}
Let's assume we have a dataset with samples $x \in X$ and labels $y \in \{-1, (N-1)\}^N$ for $N$ different 
classes sampled from a distribution $D$. If label $y$ represents class $k$, then the $k$th component 
$y^{(k)} = N-1$, all other components $h \neq k$ are  $y^{(h)} = -1$.
A feature $f$ is a function mapping that maps either to $\{0\}^N$ or to the same element in $\{0,1\}^N$.
The activation vector of a capsule $v \in \mathbb{R}^M$ of dimensionality $M$
satisfies $0 \leq ||v|| \leq 1$. We call $||v||$ the activation of a capsule which represents the probability of a feature being present or absent in the current input. 
A capsule is inactive iff $||v|| = 0$, otherwise it is (at least to some extent) active.
Every capsule $i$ of a lower level layer connects to an upper level capsule $j$ by means of the coupling coefficient 
$c_{ij}$, satisfying $\sum\limits_j c_{ij} = 1$. A large value of $c_{ij}$ indicates a strong coupling between capsules.

\begin{definition}{($\rho$-useful feature)}
   A feature $f$ is $\rho$-useful $(\rho>0)$ if it is positively correlated with the expected value of the correlation between the true label $y$ and its feature $f$:
   \[\mathbb{E}_{(x,y)\sim{}D}\left[\sum\limits^N_{n=0} y^{(n)} f(x)^{(n)}\right] \geq \rho\]

\end{definition}
In this definition we do not restrict $\rho$-useful features to only be useful for a single class, 
as features used by hidden capsules can be shared among multiple classes. 
An example of a capsule that is useful for $5$ different classes out of $10$ is shown in \cref{fig:feature_samples}.
However, a feature $g$ that is shared by all classes ($g: X \rightarrow \{1\}^N$) can never be $\rho$-useful, because: 
\[\mathbb{E}_{(x,y)\sim{}D}\left[\sum\limits^N_{n=0} y^{(n)} g(x)^{(n)}\right] = 0\]

\begin{definition}{($\gamma$-robustly useful features)}
   Given a $\rho$-useful feature $f$ with $\rho > 0$,
   $f$ is also a $\gamma$-robustly useful feature if it remains useful under some set of 
   valid adversarial perturbations \cite{adversarial-examples-are-not-bugs} $\Delta(x)$ 
   for some $\gamma > 0$:
      \[\mathbb{E}_{(x,y)\sim{}D}\left[\inf\limits_{\delta \in \Delta(x)}{\sum\limits^N_{n=0} y^{(n)} f(x + \delta)^{(n)}}\right] \geq \gamma\]
\end{definition}

\begin{definition}{(Non-robust useful feature)}
   We call a feature a non-robust useful feature if it is $\rho$-useful but not 
   $\gamma$-robustly useful. 
\end{definition}

\citet{adversarial-examples-are-not-bugs} showed that non-robust useful features are highly predictive for a class but incomprehensible for humans.
With the following definition, we ensure that a $\gamma$-capsule is only active 
if the input feature is $\gamma$-robustly useful in order to exclude incomprehensible features from being encoded by $\gamma$-capsules. 
\begin{definition}{($\gamma$-capsule)}\label{def:gamma_capsule}
   A capsule with activation $v$ is called a $\gamma$-capsule if there exists 
   a corresponding $\gamma$-robustly useful feature $f$ such that:
   \[||v_i|| > 0 \text{ iff } \sum\limits^N_n y^{(n)} f(x)^{(n)} > 0\]
\end{definition}

This definition has several implications for $\gamma$-capsules:
First, a $\gamma$-capsule can only be active if its correlated feature $f$ is positive on input $x$. Therefore, we can 
generate inputs $x$ that activate a $\gamma$-capsule to explore  the corresponding feature $f$.
The feature $f$ is $\gamma$-robustly useful and we will show in the experimental section that those features are also human comprehensible in contradiction to non-robust useful features \cite{adversarial-examples-are-not-bugs}. 
Second, we can analyze the class probabilities of a $\gamma$-capsule network for 
the generated input to determine for which classes a feature $f$ is useful. An example is given in \cref{fig:feature_samples}
where we can observe that the lower level feature that is represented by this hidden capsule
is useful for $5$ output classes.

\begin{definition}{($\gamma$-capsule network)}\label{def:gamma_capsule_network}
   Each capsule of the network satisfies \cref{def:gamma_capsule} and
   every active lower level $\gamma$-capsule $i$ selects a single upper level 
   capsule $j$ as its parent during inference:
      \[\forall k \neq j, c_{ij} > c_{ik}\]
\end{definition}
With this definition we ensure that each lower level capsule of the network is a $\gamma$-capsule. We also ensure that 
each lower level capsule selects only a single capsule to be its parent such that capsules of different layers carve a \textit{parse-tree} out of the network. The tree structure is intended to represent a hierarchical composition of objects that are made out from their components or from smaller objects. Such a parse-tree allows a solution to the problem of assigning parts to wholes \cite{transforming-autoencoders}. 
Note that 
the closer the values $c_{ij}$ of a lower level capsule $i$ are to 
$\frac{1}{J}$ (known as uniform coupling or fully connected), the weaker is the tree structure.

\subsection{Metrics}
We introduce here different metrics to determine whether a capsule network 
fulfills \cref{def:gamma_capsule} and \cref{def:gamma_capsule_network}. We group our metrics into
\textit{representation metrics}, which are aimed at evaluating whether a capsule represents a 
$\gamma$-robustly useful feature and \textit{structural metrics}, which measure how close the structure of the network is to a tree and how this tree adapts to different inputs.

\textbf{Structural metric} - $\gamma$-capsule networks should form a parse-tree structure as in \cref{def:gamma_capsule_network}. This 
structure should adapt to changing inputs because a capsule should be active only if the 
corresponding input feature is present and inactive otherwise. To measure how close the coupling of capsules is to a tree structure we introduce the \emph{T}-score. With this score we measure whether the coupling 
between capsules of different layers is close to uniform or not. For this reason, the \emph{T}-score is based on evaluating the average entropy.

For $I$ capsules in layer $l$ and $J$ capsules in layer $l+1$,
the average entropy of the coupling for a mini-batch with $M$ training examples for a single layer can be calculated as
\begin{equation}\label{eq:metric_cij_entropy}
 H_{\avg} = \frac{1}{MI} \sum^M_{m=1} \sum^I_{i=1} \sum^J_{j=1} -c^m_{ij} \log c^m_{ij}
\end{equation}
where $c^m_{ij}$ is the coupling in example $m$ from the lower level 
capsule $i$ to the upper level capsule $j$.

The value of $H_{\avg}$ changes with the number of upper level capsules $J$.
This can be prevented by normalizing the entropy using the maximum
possible entropy $\log J$ where $J$ is the number of capsules in layer
$l$. The normalized metric is shown in
\cref{eq:metric_parse_tree} and is close to $1$ whenever a parse tree
is created and close to $0$ whenever capsules are uniformly coupled.

\begin{equation} \label{eq:metric_parse_tree}
  T = 1 - \frac{H_{\avg}}{\log J}
\end{equation}

To measure whether the activation of capsules changes according to its input, we introduce 
the \emph{D}-score. For this metric, we evaluate the standard deviation 
of each capsule among the different input images.
We report the maximum standard deviation for all the 
capsules of a layer, because at least one capsule should adapt its activation 
between the different classes. We avoid using the mean or median 
since it would be sufficient a change in activation of only a few capsules 
among different input examples to obtain a good classifier.

For $I$ capsules in layer $l$ 
the D-score for a mini-batch with $M$ training examples can be calculated as
\begin{equation}\label{eq:metric_dynamicity}
   \begin{gathered}
     D = \max\limits_j^J \sqrt{\frac{1}{M}\sum\limits_m^M{\left(v_j^m - \overline{v}_j\right)^2}}
   \end{gathered}
 \end{equation}
where $v^m_j$ is the activation of capsule $j$ and input example $m$ 
and $\overline{v}_j = \mean\limits_m^M(v_j^m)$.

The D-score should be large 
whenever classes of inputs are different (i.e. different features are used for classification) 
and should be small whenever the inputs are 
of the same class (i.e. the same features are used for classification). Therefore, in the experimental section we evaluate the D-score 
using shuffled inputs from all classes and inputs that are restricted to only one class.

\textbf{Representation metric} - A $\gamma$-capsule network should represent a 
$\gamma$-robustly useful feature that is comprehensible for humans. To evaluate this property, we propose two 
different ways: 
(1) A \emph{quantitative evaluation} where we evaluate
the adversarial robustness of the network since $\gamma$-robustly useful 
features should be robust under attacks. We use 
\emph{projective gradient descent (PGD)} attacks for this evaluation, 
because \citet{resistant-to-adversraial-attacks} claims that
no other first-order adversary will find a local maxima that is significantly larger than the maxima found by PGD.
Note: For further analysis and to strengthen this argument, 
results acquired using the \emph{fast gradient sign method (FGSM)} are included in the 
supplementary material.
(2) A \emph{qualitative evaluation} to evaluate $\gamma$-capsules is to generate 
input images that activate this unit to check whether $\gamma$-capsules are comprehensible. We generate those images without 
any natural image constraint as it is generally done \citep{deep-nn-easily-fooled, deep-inside, object-detectors-emerge} since a $\gamma$-capsule should be 
only active if the input is human comprehensible. For 
example, if we generate images for output capsules, we expect those images to 
look similar to the data in the training set. The method to generate images is 
described next.

For an image $x_i$ with height $M$ and width $N$ containing randomly sampled values 
we calculate the activation loss $J$ of a single capsule $i$ with activation vector $v_i$ with
\begin{align}\label{eq:generate_images}
   J(x_i) &= \left(||v_i|| - 1\right)^2 + \lambda \sum\limits_m^M \sum\limits_n^N x_i^{m,n}
\end{align}
$x_i$ is updated iteratively with step size $\epsilon$ such that the activation loss 
$J$ is decreased:
\begin{align}
   x_i &= x_i - \epsilon \sign \left[\nabla J(x_i) \right]
\end{align}

The left term of \cref{eq:generate_images} ensures that a given capsule $i$ is activated.
The right side of the term 
ensures that pixels are only activated if they influence 
the activation of a capsule. This term is also scaled by $0 < \lambda < 1$ in order not to overcome the total loss.
Experimentally we have found that this regularization 
is only important for hidden capsules, because a capsule that represents a part of an object 
should not be influenced by other parts. For output capsules this regularization term
is not needed. We execute this process $60$ times for different random inputs to  
avoid local minima during the generation process and report the average of $5$ generated images
with the smallest $J_i(x_i)$ value. By reporting the average of multiple images 
we evaluate whether a capsule represents only one feature or multiple different 
features because in the latter case, the generated images would look blurred.

\section{Implementation of $\gamma$-capsule networks}\label{sec:implementation}
In the experimental section we show that the two predominantly 
used routing algorithms, EM-routing \cite{matrix-capsules} and routing-by-agreement (RBA) \cite{dynamic-routing} are not fitted for $\gamma$-capsule networks.
Therefore, we developed a new routing algorithm designed to 
satisfy the definitions for $\gamma$-capsule networks presented in \cref{sec:framework}. 
This new routing will ensure the required structure of $\gamma$-capsule networks (\cref{sec:implementation_routing}).
We will show next how the network can be trained to satisfy \cref{def:gamma_capsule} (\cref{sec:implementation_training}).

\subsection{A routing algorithm for $\gamma$-capsule networks}\label{sec:implementation_routing}
\begin{algorithm}[t]
   \caption{Scaled-distance-agreement (SDA) routing algorithm.  \newline $\forall$ capsules $i$ in layer $l$ 
   with $I$ capsules and $j$ in layer $l+1$ with $J$ capsules, $r$ routing iterations and 
   predictions $\hat{u}_{j|i}$ from lower level capsule activations $v_i$}\label{algo:sda} 
   \begin{algorithmic}[1]
     \Procedure{SdaRouting}{$v_i$,$\hat{u}_{j|i},r,l$}  
     \State $b_{ij} \gets 0$
     \State $\hat{u}_{j|i} \gets \min(||v_i||, ||\hat{u}_{j|i}||) \frac{\hat{u}_{j|i}}{||\hat{u}_{j|i}||} $ \label{algo:sda_line:restrict}
     \For{$r \text{ iterations}$}
       \State $c_{ij} \gets \frac{\exp(b_{ij})}{\sum_k \exp(b_{ik})}$ \label{algo:sda:line:cij}
       \State $s_j \gets \sum_{i} c_{ij}\hat{u}_{j|i}$ \label{algo:sda:line:sj} 
       \State $v_j \gets \frac{||s_j||^2}{1+||s_j||^2} \frac{s_j}{||s_j||}$ \label{algo:sda:line:vj} 
       \State $t_i \gets \frac {\log(0.9 (J-1)) - \log(1 - 0.9)} {-0.5 \mean^J_j(||\hat{u}_{j|i} - v_j||)}$ \label{algo:sda:line:t} 
       \State $b_{ij} \gets ||\hat{u}_{j|i} - v_j|| t_i$\label{algo:sda:line:bij} 
     \EndFor
     \EndProcedure
   \end{algorithmic}
 \end{algorithm}

The tree structure defined in \cref{sec:framework} is produced by the routing algorithm during inference to solve the problem of assigning parts to wholes. Therefore active lower level capsules (parts) should agree with each other to activate upper level capsules (wholes). For RBA we found that an active lower level capsule can also couple with inactive upper level capsule. For EM-routing upper level capsules can be fully active without considering the lower level capsule agreements \citep{em-routing-pitfalls}. Therefore both algorithms are not fitted to produce $\gamma$-capsule networks. A detailed analysis is provided in the supplementary material. To produce the required structure of \cref{def:gamma_capsule_network} we build our routing algorithm on RBA. Unfortunately we do not base our new routing algorithm in EM-routing as it has different pitfalls as shown by \citet{em-routing-pitfalls}.
We call our new algorithm \emph{scaled-distance-agreement} (SDA) routing algorithm.
We use inverse distances instead of the dot product to avoid that active lower level capsules couple with inactive upper level capsules. This ensures that the agreement is small if the activation of the lower level capsule is 
different than the activation of the upper level capsule. We also restrict prediction vectors to the activation of its predicting capsule as shown in
\cref{algo:sda_line:restrict} of \cref{algo:sda}. With this restriction we ensure that
a lower level capsule can activate an upper level capsule if and only if the correlated 
feature of this capsule is present in the current input. Activation and prediction vectors 
are contained within an hypersphere of radius $1$ because the maximum length of both vectors 
is $1$. The maximum possible distance between both vectors is therefore 
$2$ whenever $\hat{u}_{j|i} = -v_j$ and $||\hat{u}_{j|i}|| = ||v_j|| = 1$. So, the maximum possible coupling 
coefficient for the parent capsule will be reached whenever 
$||\hat{u}_{j|i}-v_j||=0$ for the parent capsule and
$||\hat{u}_{j|i}-v_j||=2$ for all other capsules.
The maximum coupling coefficient that is possible for the parent capsule 
with \eg $10$ upper level capsules is therefore $0.45$, because $c_{ij} = \frac{\exp(b_{ij})}{\sum_k \exp(b_{ik})} = \frac{\exp(0)}{9 \cdot \exp(-2) + \exp(0)} = 0.45$ . 
This maximum coupling coefficient gets smaller as the number of upper level 
layer capsules increases. For $128$ capsules, the maximum possible value 
for $c_{ij}$ for the parent capsule decreases to 
$c_{ij} = \frac{\exp(0)}{127 \cdot \exp(-2) + \exp(0)} = 0.05$. 

To be able to represent a strong parse tree structure with a large coupling 
coefficient for the parent capsule, we first multiply 
the distance by a scale factor $t$ to allow larger 
coupling coefficients for the parent capsule. This factor is 
calculated so that the parent capsule couples
with probability $c_{ip}$ whenever the euclidean distance of 
the parent prediction is $d_p$ and the distance to all other capsules is 
$d_o$ where $d_p < d_o$. The coupling coefficient is 
calculated using the \emph{softmax} function. Therefore, $c_{ip}$ satisfies

\begin{equation}\label{eq:sensitivity_softmax}
  c_{ip} = \frac{\exp(b_{ij})}{\sum_k^J \exp(b_{ik})} = \frac{\exp(d_p t)}{\sum^{J-1} \exp(d_o t) + \exp(d_p t)}
\end{equation}
where $J$ is the number of parent capsules. By rewriting 
\cref{eq:sensitivity_softmax} we can calculate the scale factor $t$ with

\begin{equation}\label{eq:sensitivity}
  t = \frac{\log\left(c_{ip}\left(J-1\right)\right) - \log\left(1-c_{ip}\right)}{d_p - d_o}
\end{equation}
The complete derivation of $t$ is given in the supplementary material.
Note that the scaling factor $t$ is negative ($d_p < d_o$) so that 
small distances produce a large agreement and large distances a small agreement.
Empirically we found that setting $c_{ip} = 0.9$ whenever 
$d_p = \frac{d_o}{2}$ where $d_o \approx \mean(||\hat{u}_{j|i} - v_j||)$ produces 
a strong coupling to the parent capsule. The calculation of the agreement 
using this scaled distance is shown in \cref{algo:sda:line:t} and
\cref{algo:sda:line:bij} of \cref{algo:sda}. In the experimental section 
we will show that this algorithm increases the metrics that we introduced
in \cref{sec:framework} and ensures the structure needed to satisfy 
\cref{def:gamma_capsule_network}.

\subsection{Training $\gamma$-capsule networks}\label{sec:implementation_training}
Our routing algorithm ensures the single parent constraint, such that 
a $\gamma$-capsule network represents a tree structure during inference.
SDA-routing also ensures that capsules are only active if lower level capsule votes 
agree with the general agreement. The routing does not ensure that 
each $\gamma$-capsule should represent a $\gamma$-robustly feature. 
We will train the network to minimize the empirical risk (ERM) under attack \citep{resistant-to-adversraial-attacks}
\begin{align}
   \min\limits_\theta \mathbb{E}_{(x,y)\sim{D}}\left[ \max\limits_{\delta\in\Delta(x)} L(\theta, x + \delta, y)  \right]
\end{align}
where $x$, $y$, $D$ and $\Delta$ are defined in \cref{sec:framework}.
ERM under attack has been used to train CNNs. We use ERM under attack to train capsule networks along with SDA-routing in order to obtain $\gamma$-capsule networks. 

\section{Experimental evaluation}\label{sec:evaluation}
In this section we use the framework designed in \cref{sec:framework} to 
evaluate capsule networks. First, we will evaluate the structure of capsule 
networks, after which we will analyze the features that are represented by capsules using the metrics that we introduced.
We will use MNIST from \citet{mnist}, fashionMNIST from \citet{fashion-mnist} and 
smallNorb from \citet{smallNorb} in all our experiments.

\subsection{Setup}
We will provide a comparison of matrix capsules with EM-routing, capsule networks with RBA and 
capsule networks with SDA-routing. For EM-routing we used the architecture, hyperparameters 
and implementation proposed by \citet{em-routing-pitfalls}. For 
RBA and SDA-routing we adapted the implementation from \citet{dynamic-routing} as follows: 
We added one hidden capsule layer with $32$ capsules to the CapsNet architecture.
Pixel values are normalized to the range $[0,1]$ and images are scaled to $28 \times 28$ pixels. No random data-augmentation is performed during training as our main goal is to compare the effect of the prior that we introduce in this work and 
not influence the results with any other factor. 
Details of all hyperparameters are given in the implementation that we provide on GitHub. \footnote{\url{https://github.com/peerdavid/gamma-capsule-network}}

\subsection{Evaluating structure}
In this experiment we evaluate the structure of all the proposed  networks.
The values for the T-score and the D-score (\cref{sec:framework}) are reported on the test set for a combination of all input 
classes as well as for the input class $0$. When restricting the examples to one input class we 
expect the D-score to decrease since inputs share similar features as well as the structure which is to emerge from the activated capsules, should change less. To show that the SDA-routing algorithm ensures the required structure for $\gamma$-capsule networks we minimize the empirical risk rather then the empirical risk under attack in this experiment.
In \cref{tbl:structural_experiment} we can see that RBA has a very low T-score and therefore cannot 
be fitted for $\gamma$-capsule networks. We can also 
see that the T-score for SDA is larger than the T-score for EM-routing. The D-score shows that 
neither EM-routing nor RBA adapts to the current input as it would be necessary in $\gamma$-capsule networks, since the D-score for all input classes is the same as the D-score when restricted to only one class.
For SDA-routing, the D-score that is restricted to one class is lower than the D-score 
for all classes, as expected. Therefore, neither EM-routing nor RBA are fitted to train $\gamma$-capsule networks, on the other hand, 
SDA-routing generates the required structure for $\gamma$-capsule networks.
In order to better show this difference, we provide an activation map for the first hidden capsule layer 
in the supplementary material. We also want to point out 
that RBA and SDA have a larger accuracy than EM-routing in our setup without data augmentation. RBA and SDA accuracy results are very  similar.

\begin{table}
   \begin{center}
   \begin{tabular}{ll|cc|ccc}
   \toprule
                  &         & \multicolumn{2}{c|}{[0]} &  \multicolumn{3}{c}{[0-9]}  \\
   Dataset        & Alg.    & T & D & T & D & Acc. \\
   \midrule
   MNIST          & RBA     & 0.02 & 0.09 & 0.02 & 0.11 & \bfseries 99.12 \\
                  & EM      & 0.27 & 0.08 & 0.31 & 0.08 & 98.58 \\
                  & SDA     & \bfseries 0.49 & \bfseries 0.23 & \bfseries 0.48 & \bfseries 0.42 & 98.91 \\
   \midrule
   fashion        & RBA     & 0.02  & 0.21 & 0.02 & 0.21 & 90.10 \\
   MNIST       & EM      & 0.23  & 0.17 & 0.24 & 0.15 & 88.69 \\
                  & SDA     & \bfseries 0.48  & \bfseries 0.30 & \bfseries 0.48 & \bfseries 0.40 & \bfseries 90.74 \\
   \midrule
   norb           & RBA     & 0.02 & 0.24 & 0.01 & 0.23 & \bfseries 89.82 \\
                  & EM      & 0.31 & 0.12 & 0.28 & 0.11 & 83.10 \\
                  & SDA     & \bfseries 0.47 & \bfseries 0.39 & \bfseries 0.47 & \bfseries 0.41 & 88.61 \\
   \bottomrule
   \end{tabular}
   \end{center}
   \caption{Structural metrics for different datasets using all classes [0-9] or only one class [0].}
   \label{tbl:structural_experiment}
\end{table}


\subsection{Evaluating representation}
\begin{table}[t]
   \begin{center}
   \begin{tabular}{ll|ccc}
   \toprule
                  &         & \multicolumn{3}{c}{PGD ($\epsilon$)} \\
   Dataset        & Alg.    & $\epsilon=0.1$ & $\epsilon=0.3$ & $\epsilon=0.5$ \\
   \midrule
   MNIST          & RBA     & 55.76 & 0.25 & 0.0 \\
                  & EM      & 10.49 & 0.09 & 0.0 \\
                  & SDA     & \bfseries 97.10 & \bfseries 92.40 & \bfseries 20.11 \\
   \midrule
   fashionMNIST   & RBA     & 3.14 & 0.26 & 0.08 \\
                  & EM      & 0.0 & 0.0 & 0.0 \\
                  & SDA     & \bfseries 71.63 & \bfseries 59.01 & \bfseries 1.89 \\
   \midrule
   smallNorb      & RBA     & 20.99 & 0.17 & 0.0 \\
                  & EM      & 7.09  & 0.02  & 0.0 \\
                  & SDA     & \bfseries 64.58 & \bfseries 34.89 & \bfseries 18.72 \\
   \bottomrule
   \end{tabular}
   \end{center}
   \caption{Accuracy of RBA, EM-routing and SDA-routing under attack. To attack the network with PGD 
   we used the same parameters as in \citet{resistant-to-adversraial-attacks} and varied $\epsilon$ from 
   $0.1$ to $0.5$.}
   \label{tbl:adversarial_experiment_pgd}
\end{table}

For the \emph{quantitative evaluation} we compare the adversarial robustness 
of all algorithms and all datasets using PGD. Results for the FGSM are added to the supplementary material. 
In the previous experiments we have seen that SDA-routing ensures the required structure for $\gamma$-capsule networks. In this additional experiment, we train the network using ERM under attack as described 
in \cref{sec:implementation}. We use PGD with $a=0.01$, $k=40$ and $\epsilon=0.3$ for the 
inner maximization \citep{resistant-to-adversraial-attacks}.
We can see in \cref{tbl:adversarial_experiment_pgd} that SDA-routing is much more robust against adversarial attacks than RBA and 
EM-routing. With an attack rate of $\epsilon=0.3$ SDA has an accuracy of $92.40\%$ whereas the 
accuracy of RBA and EM-routing drops to almost $0$ on MNIST. This supports our claim that the capsules from our 
network are indeed $\gamma$-capsules and they represent $\gamma$-robustly useful features, whereas capsules of EM-routing and RBA rely on useful non-robust features. 
We also want to mention that for MNIST, our method still has an accuracy of $20\%$ under strong attacks.
The CNNs that were trained by \citet{resistant-to-adversraial-attacks} have an accuracy of 
$0\%$ after $\epsilon=0.3$, even though the authors reported in the appendix adversarial images for strong attacks 
that are still recognizable.

\begin{figure*}[t]
   \begin{center}
      \includegraphics[width=0.95\textwidth]{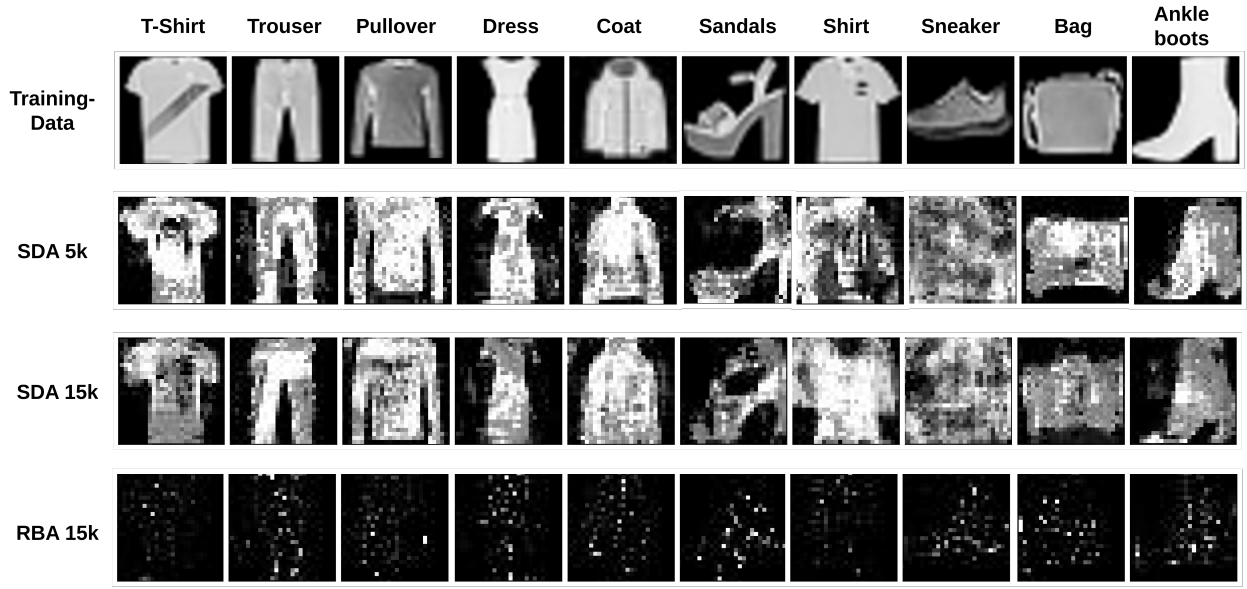}
   \end{center}
      \caption{The first row shows a random input image of each class of the training set. The other 
      rows shows images generated with the approach from \cref{sec:framework} for $\gamma$-capsule networks.}
   \label{fig:generated_images}
\end{figure*}

We now continue with a \emph{qualitative evaluation} where we analyze visually the features that are represented by the 
individual capsules of the network.
In this qualitative approach, we interpret the input features for the fashionMNIST dataset, 
because after the quantitative evaluation we have seen that this is the most challenging dataset for SDA-routing 
(see $\epsilon=0.5$ for fashionMNIST). In \cref{fig:generated_images} we show input images that are 
generated for all output capsules when applying our method
after $5$k (row $2$) and $15$k training steps (row $3$), for RBA after $15$k steps (row $4$).
One observation of this experiment is that $\gamma$-capsules are only activated if 
inputs are close to the given training data whereas for RBA, only 
separate lonely pixels are needed to activate a capsule (more detailed results for the activation are given in the supplementary material). This qualitative analysis demonstrates  
that $\gamma$-capsules are comprehensible for humans and shows why it is harder to attack 
$\gamma$-capsule networks. We now want to outline one interesting internal detail that we can 
describe because of the explainability property of $\gamma$-capsule networks: After $5$k steps (row 2) in \cref{fig:generated_images} the 
classes shirt and sneaker are not 
comprehensible for $\gamma$-capsules indicating that the $\gamma$-capsule network was not able to extract features for those classes. To further study this anomaly we obtained the confusion matrix in 
\cref{tbl:confusion_matrix} and were able to confirm that no features are found for class $6$ (shirt), 
and therefore this class is not correctly classified. On the other hand, the confusion matrix 
shows also that classification is done correctly for the sneakers class. We conclude that 
the sneakers class is used as a none-of-the-above class because it is active for arbitrary inputs
as shown in \cref{fig:generated_images}. We further analyzed this phenomenon  
by continuing the training and found that (1) features for the shirt
class are learned after $15$k steps and (2) the sneakers class is still used as none-of-the-above class. This evaluation shows that it is really hard for the network to learn features that clearly classifies shirts. Also, if we analyze the generated images after $15$k steps, features for shirt and t-shirt are very close. This is also supported by the confusion-matrix because this class is often misclassified as t-shirt (\cref{tbl:confusion_matrix}).  For the sneaker class we can see that also after $15$k steps it is still better for the network to learn a none-of-the-above class rather than features for sneakers. We leave the question when none-of-the-above classes are learned open for future work but we believe that it is often easier for the network to learn such a class than extracting real features. At this point we also want to mention that the reconstruction network only hides this none-of-the-above class and is therefore not really helpful to explain the cause of active capsules. Images of the reconstruction network are shown in the supplementary material.

\begin{table}[t]
   \begin{center}
      \scriptsize
   \begin{tabular}{c|cccccccccc}
   \toprule
     & 0 & 1 & 2 & 3 & 4 & 5 & 6 & 7 & 8 & 9 \\ \midrule
   0 & \textbf{805} &  32 &  62 &  63 &  12 &   7 &   \textbf{0} &   0 &  39 &   2 \\ 
   1 & 15 & \textbf{907} &   6 &  42 &   3 &   0 &   \textbf{0} &   0 &   0 &   0 \\ 
   2 & 36 &  12 & \textbf{622} &  25 & 336 &   0 &   \textbf{0} &   0 &  39 &   0 \\
   3 & 48 &  27 &  15 & \textbf{796} &  67 &   0 &   \textbf{0} &   0 &  15 &   3 \\ 
   4 & 10 &  15 &  98 &  36 & \textbf{877} &   0 &   \textbf{0} &   0 &   8 &   3 \\ 
   5 &  0 &   0 &   0 &   0 &   0 & \textbf{779} &   \textbf{0} & 116 &   4 &  60 \\ 
   6 &327 &  13 & 204 &  31 & 369 &   3 &   \textbf{0} &   0 &  50 &   0 \\
   7 &  0 &   0 &   0 &   0 &   0 &  45 &   \textbf{0} & \textbf{760} &   0 & 184 \\ 
   8 &  4 &   0 &   8 &   9 &  14 &  17 &   \textbf{0} &   3 & \textbf{965} &   7 \\ 
   9 &  0 &   0 &   0 &   0 &   0 &   2 &   \textbf{0} &  14 &   0 & \textbf{929} \\ 
   \bottomrule
   \end{tabular}
   \end{center}
   \caption{Confusion matrix after $5k$ steps. The class shirt is not learned correctly.}
   \label{tbl:confusion_matrix}
\end{table}

\begin{table}[t]
   \begin{center}
   \begin{tabular}{>{\centering\arraybackslash} m{1.8cm} >{\centering\arraybackslash} m{1.8cm} >{\centering\arraybackslash} m{1.8cm}}
   \toprule
   Hidden Caps. & Image & Output class \\
   \midrule
   0 & \includegraphics[width=0.65\linewidth]{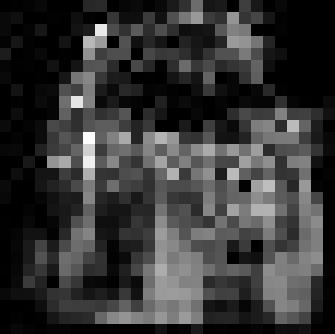} & Bag \\
   1 & \includegraphics[width=0.65\linewidth]{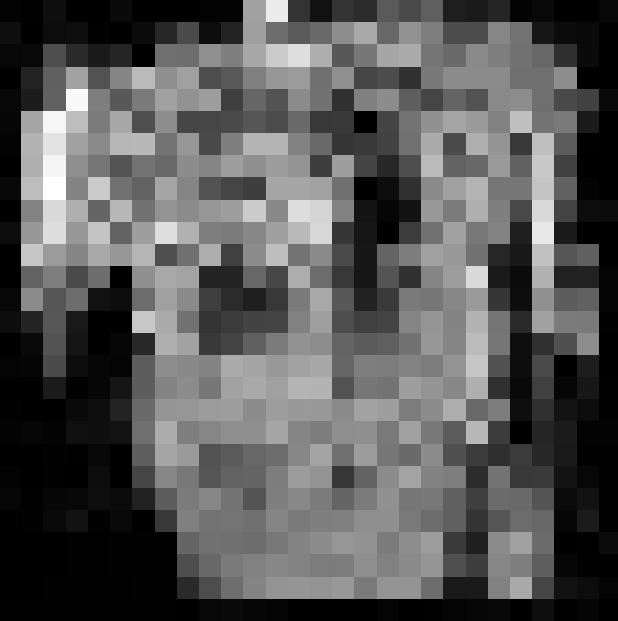} & T-Shirt \\
   6 & \includegraphics[width=0.65\linewidth]{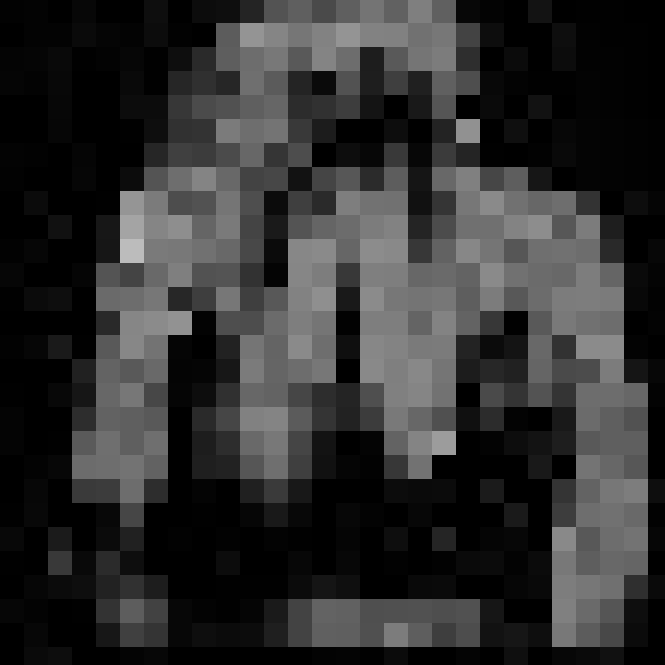} & Pullover \\
   7 & \includegraphics[width=0.65\linewidth]{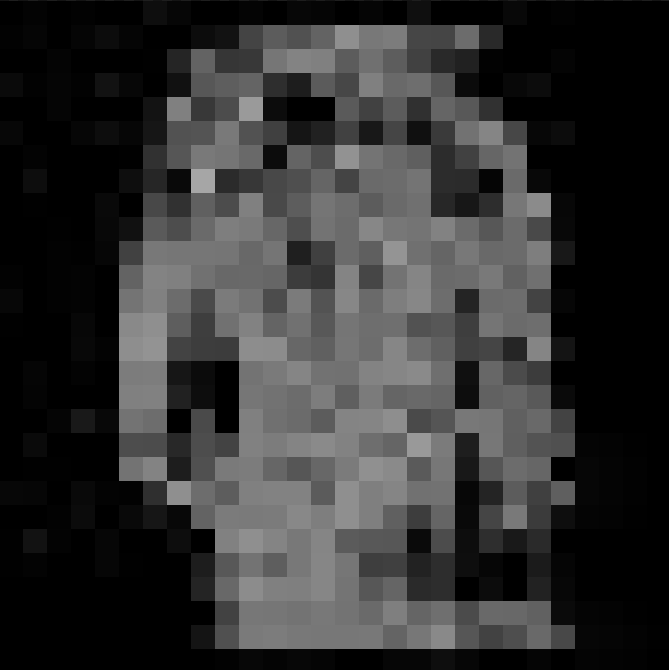} & Coat \\
   17 & \includegraphics[width=0.65\linewidth]{images/gamma_hidden_17} & Dress \\
   20 & \includegraphics[width=0.65\linewidth]{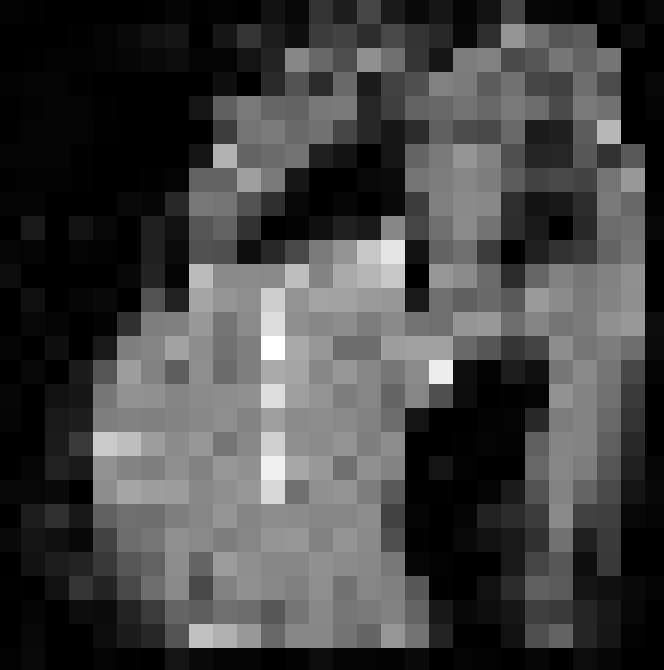} & Sandals \\
   \bottomrule
   \end{tabular}
   \end{center}
   \caption{Generated features to activate single capsules of the first hidden layer}
   \label{tbl:hidden_capsules}
\end{table}




Output capsules are comprehensible and therefore they satisfy \cref{def:gamma_capsule}.
We now show for some hidden capsules that they are also comprehensible and conclude that the network that we 
produced is a $\gamma$-capsule network. One example for a hidden capsule that is comprehensible is already 
presented in \cref{fig:feature_samples}. We claim that this capsule represents the body of different 
clothes and therefore the classes t-shirt, pullover, dress, coat and shirt are active for this input.
In \cref{tbl:hidden_capsules} we report more hidden capsules. Together with the image that is generated 
to activate the hidden unit we report the most active output class. As we can see in all cases,  
the hidden capsule is correlated with the output class and therefore the cause for an active hidden capsule becomes explainable. 
We can also see that hidden capsules represent human comprehensible objects or part of objects. For example, capsule $6$ 
encodes the upper part of the pullover and the long sleeves, features that separates it from most of the other classes. On the other hand, capsules $7$ and $17$ only encode the body 
without arms. Others, such as capsule $0$ or $1$  encode the whole object. 

\section{Discussion}\label{sec:conclusion}
In this paper we introduced a new type of capsule networks 
which we call $\gamma$-capsule networks. This type of network
implements an inductive bias that is motivated by the biological TE neurons from the inferiortemporal cortex.
Differently to previous work on capsule networks \cite{matrix-capsules, dynamic-routing},
we define this prior formally and we also provide metrics to ensure 
that the prior is fulfilled properly. We have shown that the most common 
routing algorithms for capsule networks, namely RBA \cite{dynamic-routing} and EM-routing \cite{matrix-capsules} are not fit for implementing this inductive bias. One limitation of routing-by-agreement is that the routing coefficient is calculated without considering the activation of the upper level capsules. To overcome this limitation, we introduce SDA-routing, which considers this case and we show experimentally that 
this algorithm can be used to produce $\gamma$-capsule networks. After training the network with SDA-routing using ERM under attack we have shown that $\gamma$-capsule networks are more robust than classical capsule networks. Regular capsule networks are no more robust against adversarial attacks than 
CNNs \cite{capsule-adversarial-attacks}. The robustness that is reported in this experimental work is a property exclusively of $\gamma$-capsule networks, one not necessarily present in capsule networks in general. Additionally $\gamma$-capsule networks have an even higher degree of robustness  under strong attacks than CNNs trained specifically against adversarial attacks \cite{resistant-to-adversraial-attacks}. 
The convolutional filters that are learned for $\gamma$-capsule networks contain highly concentrated weights (see supplementary material). Robust CNNs learn similar filters \cite{resistant-to-adversraial-attacks}, therefore we conclude that this robustness under strong attack is a property of the $\gamma$-capsule layers rather than the convolutional layers. As opposed to previous work \cite{detecting-adversaries}, our approach encodes directly $\gamma$-robustly useful features \cite{adversarial-examples-are-not-bugs} that are robust against attacks   
 instead of using a reconstruction network to detect the adversarials.
 We have also shown that the images of reconstruction networks not necessarily represent the real cause for an active capsule. In other words, reconstruction networks can not be used to explain features that activate single capsules. On the other hand, for $\gamma$-capsule networks this is possible because human comprehensible images are generated without the use of reconstruction networks and natural image constraint \cite{deep-nn-easily-fooled, deep-inside, object-detectors-emerge} that possible hides important internals.
We conclude that $\gamma$-capsule networks can be of interest, since (1) it can be very challenging to succeed in an attack against $\gamma$-capsule networks and (2) an interpretation can be directly made of what $\gamma$-capsules are encoding without hiding any important details.

{\small
\bibliographystyle{plainnat}
\bibliography{egbib}

\begin{thebibliography}{25}
\providecommand{\natexlab}[1]{#1}
\providecommand{\url}[1]{\texttt{#1}}
\expandafter\ifx\csname urlstyle\endcsname\relax
  \providecommand{\doi}[1]{doi: #1}\else
  \providecommand{\doi}{doi: \begingroup \urlstyle{rm}\Url}\fi

\bibitem[(anonymous)(2017)]{em-routing-beta}
(anonymous).
\newblock {beta\_v and beta\_a}.
\newblock \url{https://openreview.net/forum?id=HJWLfGWRb&noteId=ryTPZJd-f},
  2017.
\newblock [Online; accessed 11/2019].

\bibitem[Duarte et~al.(2018)Duarte, Rawat, and Shah]{video-capsnet}
Kevin Duarte, Yogesh Rawat, and Mubarak Shah.
\newblock Videocapsulenet: A simplified network for action detection.
\newblock In \emph{Advances in Neural Information Processing Systems}, pages
  7610--7619, 2018.

\bibitem[Fukushima(1980)]{neocognitron-self-org-nn}
Kunihiko Fukushima.
\newblock Neocognitron: A self-organizing neural network model for a mechanism
  of pattern recognition unaffected by shift in position.
\newblock \emph{Biological Cybernetics}, 36\penalty0 (4):\penalty0 193--202,
  Apr 1980.
\newblock ISSN 1432-0770.
\newblock \doi{10.1007/BF00344251}.

\bibitem[Gritzman(2019)]{em-routing-pitfalls}
Ashley~Daniel Gritzman.
\newblock Avoiding implementation pitfalls of ``matrix capsules with em
  routing'' by hinton et al.
\newblock In An~Zeng, Dan Pan, Tianyong Hao, Daoqiang Zhang, Yiyu Shi, and
  Xiaowei Song, editors, \emph{Human Brain and Artificial Intelligence}, pages
  224--234, Singapore, 2019. Springer Singapore.
\newblock ISBN 978-981-15-1398-5.

\bibitem[Hinton et~al.(2018)Hinton, Sabour, and Frosst]{matrix-capsules}
Geoffrey Hinton, Sara Sabour, and Nicholas Frosst.
\newblock Matrix capsules with em routing.
\newblock In \emph{6th International Conference on Learning Representations,
  ICLR}, 2018.

\bibitem[Hinton et~al.(2011)Hinton, Krizhevsky, and
  Wang]{transforming-autoencoders}
Geoffrey~E Hinton, Alex Krizhevsky, and Sida~D Wang.
\newblock Transforming auto-encoders.
\newblock In \emph{International Conference on Artificial Neural Networks},
  pages 44--51. Springer, 2011.

\bibitem[Ilyas et~al.(2019)Ilyas, Santurkar, Tsipras, Engstrom, Tran, and
  Madry]{adversarial-examples-are-not-bugs}
Andrew Ilyas, Shibani Santurkar, Dimitris Tsipras, Logan Engstrom, Brandon
  Tran, and Aleksander Madry.
\newblock Adversarial examples are not bugs, they are features.
\newblock In \emph{NeurIPS}, 2019.

\bibitem[Kosiorek et~al.(2019)Kosiorek, Sabour, Teh, and
  Hinton]{stacked-capsule-autoencoders}
Adam~R Kosiorek, Sara Sabour, Yee~Whye Teh, and Geoffrey~E Hinton.
\newblock Stacked capsule autoencoders.
\newblock \emph{NeurIPS}, 2019.

\bibitem[Lab(2019)]{tesla-autopilot}
Tencent Keen~Security Lab.
\newblock Experimental security research of tesla autopilot., 2019.

\bibitem[LeCun and Cortes(2010)]{mnist}
Yann LeCun and Corinna Cortes.
\newblock {MNIST} handwritten digit database.
\newblock 2010.
\newblock URL \url{http://yann.lecun.com/exdb/mnist/}.

\bibitem[LeCun et~al.(1999)LeCun, Haffner, Bottou, and Bengio]{cnn}
Yann LeCun, Patrick Haffner, Léon Bottou, and Yoshua Bengio.
\newblock Object recognition with gradient-based learning.
\newblock In \emph{Shape, Contour and Grouping in Computer Vision}, volume 1681
  of \emph{Lecture Notes in Computer Science}, pages 319--. Springer, 1999.

\bibitem[LeCun et~al.(2004)LeCun, Huang, Bottou, et~al.]{smallNorb}
Yann LeCun, Fu~Jie Huang, Leon Bottou, et~al.
\newblock Learning methods for generic object recognition with invariance to
  pose and lighting.
\newblock In \emph{CVPR (2)}, pages 97--104. Citeseer, 2004.

\bibitem[Madry et~al.(2018)Madry, Makelov, Schmidt, Tsipras, and
  Vladu]{resistant-to-adversraial-attacks}
Aleksander Madry, Aleksandar Makelov, Ludwig Schmidt, Dimitris Tsipras, and
  Adrian Vladu.
\newblock Towards deep learning models resistant to adversarial attacks.
\newblock In \emph{ICLR (Poster)}, 2018.

\bibitem[Michels et~al.(2019)Michels, Uelwer, Upschulte, and
  Harmeling]{capsule-adversarial-attacks}
Felix Michels, Tobias Uelwer, Eric Upschulte, and Stefan Harmeling.
\newblock On the vulnerability of capsule networks to adversarial attacks.
\newblock In \emph{ICML 2019 Workshop on Security and Privacy of Machine
  Learning}, 2019.

\bibitem[Mobiny and Van~Nguyen(2018)]{lung-cancer-screening}
Aryan Mobiny and Hien Van~Nguyen.
\newblock Fast capsnet for lung cancer screening.
\newblock In \emph{International Conference on Medical Image Computing and
  Computer-Assisted Intervention}, pages 741--749. Springer, 2018.

\bibitem[Nguyen et~al.(2015)Nguyen, Yosinski, and Clune]{deep-nn-easily-fooled}
Anh Nguyen, Jason Yosinski, and Jeff Clune.
\newblock Deep neural networks are easily fooled: High confidence predictions
  for unrecognizable images.
\newblock In \emph{Proceedings of the IEEE conference on computer vision and
  pattern recognition}, pages 427--436, 2015.

\bibitem[Qin et~al.(2019)Qin, Frosst, Sabour, Raffel, Cottrell, and
  Hinton]{detecting-adversaries}
Yao Qin, Nicholas Frosst, Sara Sabour, Colin Raffel, Garrison~W. Cottrell, and
  Geoffrey~E. Hinton.
\newblock Detecting and diagnosing adversarial images with class-conditional
  capsule reconstructions.
\newblock \emph{CoRR}, abs/1907.02957, 2019.
\newblock URL \url{http://arxiv.org/abs/1907.02957}.

\bibitem[Rajasegaran et~al.(2019)Rajasegaran, Jayasundara, Jayasekara,
  Jayasekara, Seneviratne, and Rodrigo]{deepcaps}
Jathushan Rajasegaran, Vinoj Jayasundara, Sandaru Jayasekara, Hirunima
  Jayasekara, Suranga Seneviratne, and Ranga Rodrigo.
\newblock Deepcaps: Going deeper with capsule networks.
\newblock In \emph{Proceedings of the IEEE Conference on Computer Vision and
  Pattern Recognition}, pages 10725--10733, 2019.

\bibitem[Sabour et~al.(2017)Sabour, Frosst, and Hinton]{dynamic-routing}
Sara Sabour, Nicholas Frosst, and Geoffrey~E Hinton.
\newblock Dynamic routing between capsules.
\newblock In \emph{Advances in neural information processing systems}, pages
  3856--3866, 2017.

\bibitem[Simonyan et~al.(2014)Simonyan, Vedaldi, and Zisserman]{deep-inside}
Karen Simonyan, Andrea Vedaldi, and Andrew Zisserman.
\newblock Deep inside convolutional networks: Visualising image classification
  models and saliency maps.
\newblock \emph{ICLR workshop}, 2014.

\bibitem[Tanaka(1996)]{inferotemporal-cortex}
K.~Tanaka.
\newblock Inferotemporal cortex and object vision.
\newblock \emph{Annual Review of Neuroscience}, 19:\penalty0 109--139, 1996.

\bibitem[Xiao et~al.(2017)Xiao, Rasul, and Vollgraf]{fashion-mnist}
Han Xiao, Kashif Rasul, and Roland Vollgraf.
\newblock Fashion-mnist: a novel image dataset for benchmarking machine
  learning algorithms, 2017.

\bibitem[Zador(2019)]{critique-pure-learning}
Anthony Zador.
\newblock A critique of pure learning and what artificial neural networks can
  learn from animal brains.
\newblock \emph{Nature Communications}, 10, 12 2019.

\bibitem[Zhao et~al.(2019)Zhao, Birdal, Deng, and Tombari]{3d-point-capsules}
Yongheng Zhao, Tolga Birdal, Haowen Deng, and Federico Tombari.
\newblock 3d point capsule networks.
\newblock In \emph{Proceedings of the IEEE Conference on Computer Vision and
  Pattern Recognition}, pages 1009--1018, 2019.

\bibitem[Zhou et~al.(2015)Zhou, Khosla, Lapedriza, Oliva, and
  Torralba]{object-detectors-emerge}
Bolei Zhou, Aditya Khosla, Agata Lapedriza, Aude Oliva, and Antonio Torralba.
\newblock Object detectors emerge in deep scene cnns.
\newblock In \emph{International Conference on Learning Representations
  (ICLR)}, 2015.

\end{thebibliography}
}

\appendix
\section{Scale parameter $t$ of SDA routing}
In this section we show how to derive the scale factor $t$ from
$c_{ip} = \frac{\exp(d_p t)}{\sum^{J-1} \exp(d_o t) + \exp(d_p t)}$.
$c_{ip}$ is the coupling coefficient for the parent capsule whenever the distance to the parent capsule is $d_p$ and the 
distance to all other remaining capsules is $d_o$ for $J$ capsules 
in the upper layer. Therefore:

\begin{align*}
    && c_{ip} = \frac{\exp(b_{ij})}{\sum_k^J \exp(b_{ik})} = \frac{\exp(d_p t)}{\sum^{J-1} \exp(d_o t) + \exp(d_p t)} \\
    \iff && c_{ip} = \frac{\exp(d_p t)}{(J-1) \exp(d_o t) + \exp(d_p t)} \\
    \iff && c_{ip} (J-1) \exp(d_o t) + c_{ip} \exp(d_p t) = \exp(d_p t) \\
    \iff && c_{ip} (J-1) \exp(d_o t) = (1 - c_{ip}) \exp(d_p t) \\
    \iff && \log(c_{ip} (J-1) \exp(d_o t)) = \log((1 - c_{ip}) \exp(d_p t)) \\
    \iff && d_o t + \log(c_{ip} (J-1)) = d_p t + \log(1 - c_{ip}) \\
    \iff && \log(c_{ip} (J-1)) - \log(1 - c_{ip}) = d_p t - d_o t \\
    \iff && \frac{\log(c_{ip} (J-1)) - \log(1 - c_{ip})}{d_p - d_o} = t \\
\end{align*}

We can see that this function is well defined iff $J > 1$. This is not a limitation because a capsule should represent only \emph{one} object (or \emph{one} part of an object). Therefore also for binary classification two output capsules should be used rather than one.

\section{Pseudo code for sampling algorithm}
In this section we give the pseudo code to generate input features that activate single units of the $\gamma$-capsule network. Images are initially sampled from noise and not from correctly classified natural images like it is done in related work. Additionally no natural image constraint is used to generate input features \cite{deep-nn-easily-fooled, deep-inside, object-detectors-emerge}:

\begin{algorithm}[H]
    \caption{Algorithm to generate input features that activate a single capsule $i$ with activation $||v_i||$ for images of height $M$ and width $N$} \label{alg:generate}
    \begin{algorithmic}[1]
        \Procedure{GenerateInputFeatures}{}  
        \State $\epsilon \gets 0.01$
        \State $\lambda \gets 10^{-5}$
        \State $arr \gets \text{empty array}$
        \For{$60 \text{ iterations}$} \Comment{Produce $60$ different inputs}
            \State Create image $x_i$ with random pixel values
            \For{$1000 \text{ iterations}$}
                \State Get activation $||v_i||$ for input $x_i$
                \State $J(x_i) = \left(||v_i|| - 1\right)^2 + \lambda \sum\limits_m^M \sum\limits_n^N x_i^{m,n}$
                \State $x_i \gets x_i - \epsilon \sign \left[\nabla J(x_i) \right]$
            \EndFor
            \State Insert $x_i$ into $arr$
        \EndFor
        
        \State $x_i \gets$ average of $x \in arr$ for $5$ smallest $J(x)$
        \State \Return $x_i$
        \EndProcedure
    \end{algorithmic}
 \end{algorithm}

\section{RBA and EM routing for $\gamma$-capsule networks}
For $\gamma$-capsule networks it is important that capsules are only active if the correlated feature exists in the current input, because we want to evaluate the input features that activate capsules. Therefore upper level capsules should be produced by active lower level capsules to solve the problem of assigning parts to wholes. We found the following problems in existing routing algorithms:

\textbf{RBA: }
The log prior that is used to calculate the coupling coefficient $c_{j}$ is calculated with $b_{ij} = b_{ij} + ||v_j|| ||\hat{u}_{j|i}|| \cos{\alpha}$. If we assume that $\alpha \approx 0$ and the activation $||v_j||$ is small, then a large coupling can simply be produced if the vote $||\hat{u}_{j|i}||$ is large. Note that $\hat{u}_{j|i} = W_{ij}v_i$ with weight matrix $W_{ij}$ learned through backpropagation and therefore $ 0 \leq ||\hat{u}_{j|i}|| \leq \infty$. Therefore this algorithm must be adapted such that an active lower level capsule can not couple with an inactive upper level capsule. We have shown how this can be done with inverse distances rather than the dot product.

\textbf{EM routing: }
For EM routing we found that the activation of a capsule can be activated even when lower level capsules do not match with each other: The activation for an upper capsule $j$ is calculated with $a_j = logistic(\lambda(\beta_a - \sum_h cost_j^h))$. The value $\beta_a$ is learned through backpropagation \cite{matrix-capsules} for each capsule type individual as claimed by the original authors on OpenReview.net \cite{em-routing-beta}. Therefore also if votes from lower level capsules do not match the parent capsule (i.e. a large value for $\sum_h cost_j^h$), the capsule $j$ can be activated by simply learning a large value for $\beta_a$ (note that there is an individual $\beta_a$ for each $j$). This was already reported by \citet{em-routing-pitfalls} and they claim that a carefully initialization could help. We used the proposed method \cite{em-routing-pitfalls} but have seen experimentally (see \cref{fig:activation_map} or the reported D-score in the paper) that matrix capsules with EM routing do not adapt the activation of capsules for different inputs. Therefore EM routing is not fitted for $\gamma$-capsule networks.

\section{Activation maps of hidden capsule layers}
\begin{figure}[H]
    \begin{center}
       \includegraphics[width=0.45\textwidth]{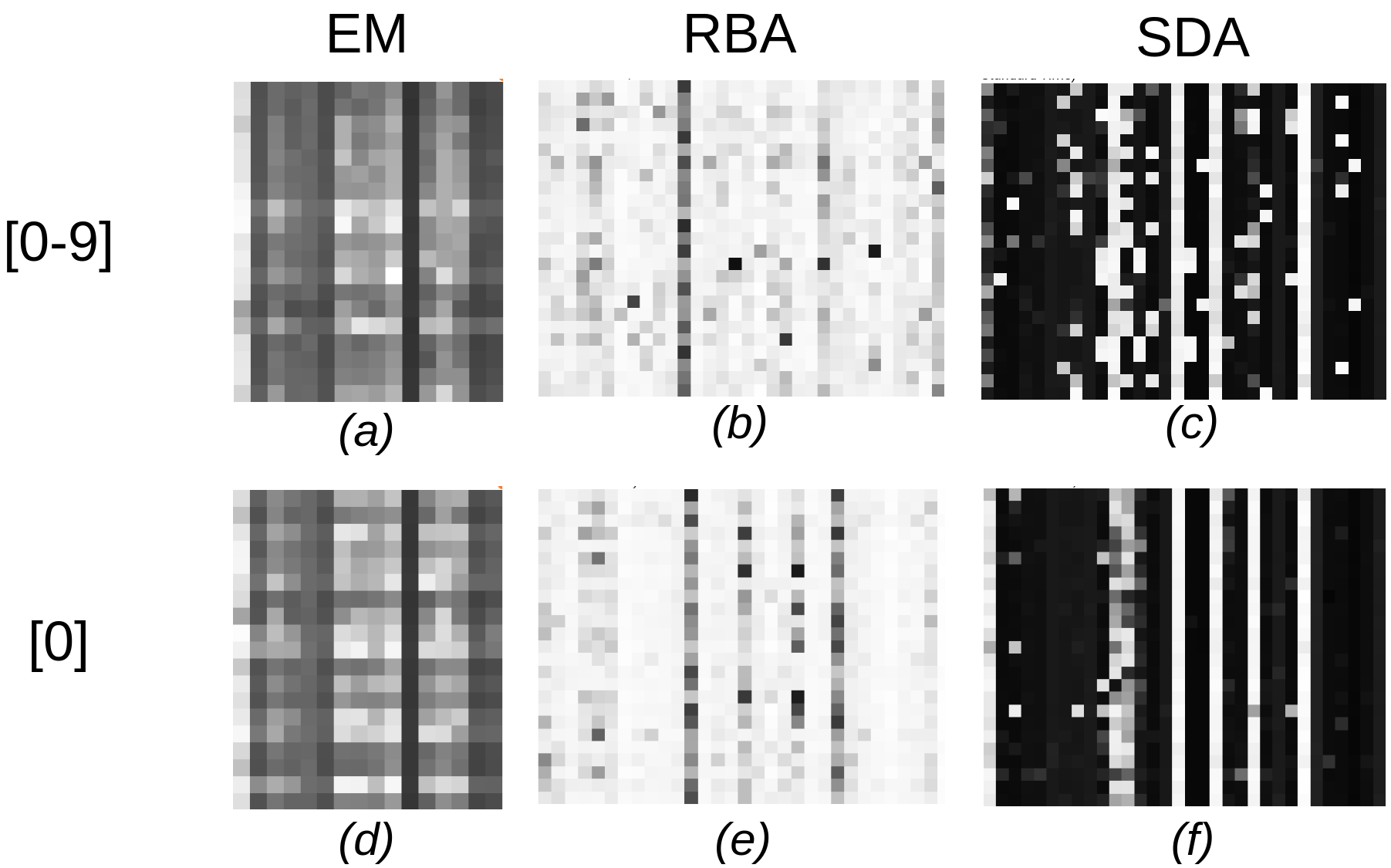}
    \end{center}
       \caption{Each pixel-column of an image shows the activation of a hidden capsule. Each pixel-row shows 
       the activation for different input examples. (a), (b) and (c) are activations of capsules for different shuffled input classes 
       whereas in (d), (e) and (f) only class $0$ is used.}
    \label{fig:activation_map}
 \end{figure}

To see the difference between different $D$-scores we additionally plot $6$ different activation maps in \cref{fig:activation_map}: For input examples that are randomly shuffled for all classes (a, b, c) and input examples restricted to only one class (d, e, f) and the $3$ algorithms (EM-routing, RBA and SDA-routing). Each single pixel-row of an image represents a different input example and each pixel-column the activation of a single capsule. For EM routing we see $16$ columns because the architecture proposed by \citet{em-routing-pitfalls} has $16$ hidden capsule types and for RBA and SDA we see the $32$ hidden capsules. 
 We can see that the activation map for EM routing and RBA routing for all classes (a, b) looks very similar to the restricted input (d, e) showing that capsules do not adapt to the input. Therefore the requirement that a $\gamma$-capsule is active iff the input feature exists in the current input is not fulfilled for EM routing and RBA. On the other hand for SDA-routing we can see that this is the case, because capsules are highly active or completely inactive (c) and if we restrict the input to one class (f) almost always the same capsules are active.

\section{Robustness against FGSM attacks}
We mentioned in the paper that we also attacked all networks with the FGSM to show that our network fulfills all requirements that are needed for $\gamma$-capsule networks. The results for FGSM are reported in \cref{tbl:adversarial_experiment_fgsm}.
\begin{table}[H]
   \begin{center}
   \begin{tabular}{ll|ccc}
   \toprule
                  &         & \multicolumn{3}{c}{FGSM} \\
   Dataset        & Alg.    & $\epsilon=0.1$ & $\epsilon=0.3$ & $\epsilon=0.5$ \\
   \midrule
   MNIST          & RBA     & 79.29 & 26.61 & 28.94 \\
                  & EM      & 74.91 & 46.59 & 22.67 \\
                  & SDA     & \bfseries 97.37 & \bfseries 94.40 & \bfseries 41.99 \\
   \hline
   fashionMNIST   & RBA     & 25.16 & 14.71 & \bfseries 18.29 \\
                  & EM      & 1.96  & 5.31  & 7.19 \\ 
                  & SDA     & \bfseries 73.49 & \bfseries 65.16 & 12.75 \\
   \hline
   smallNorb      & RBA     & 35.38 & 7.94  & 2.61 \\ 
                  & EM      & 28.38 & 15.27 & 14.37 \\
                  & SDA     & \bfseries 80.01 & \bfseries 75.77 & \bfseries 64.04 \\
   \bottomrule
   \end{tabular}
   \end{center}
   \caption{Accuracy of RBA, EM routing and SDA routing under attack. To attack the network with PGD 
   we used the same parameters as in \citet{resistant-to-adversraial-attacks} and varied $\epsilon$ from 
   $0.1$ to $0.5$.}
   \label{tbl:adversarial_experiment_fgsm}
\end{table}
We can see that SDA routing is again much more robust under attack than RBA and EM routing. We also want to mention that the FGSM attack is less successfull than the PGD attack (see paper) which supports the claim that as long as 
the adversary only uses the gradient of the loss function, the local maxima that is found by PGD is not significantly larger than other first order adversary \cite{resistant-to-adversraial-attacks}.

\section{Reconstruction of sneakers}
\begin{figure}[H]
    \centering
    \begin{subfigure}{.35\linewidth}
       \centering
       \includegraphics[width=0.8\linewidth]{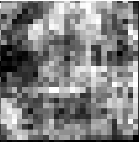}\label{fig:sneakers_reproduction_a}
       \caption{}
    \end{subfigure}
    \begin{subfigure}{.35\linewidth}
       \centering
       \includegraphics[width=0.8\linewidth]{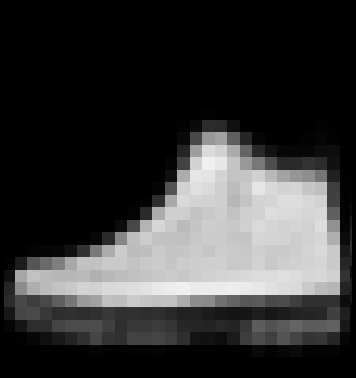}\label{fig:sneakers_reproduction_b}
       \caption{}
    \end{subfigure}

    \caption{In (a) the input image is shown that we generated with our novel method to find relevant features and in (b) an image that is reconstructed by the reconstruction network is shown.}
    \label{fig:sneakers_reproduction}
 \end{figure}

We mentioned in the paper that although we found that the sneakers class represents a \emph{none-of-the-above} class, the reconstruction network learned to reconstruct sneakers because this minimizes the reconstruction loss. An image of a sneaker that is generated with \cref{alg:generate} in comparison with the reconstruction from the reconstruction network is shown in \cref{fig:sneakers_reproduction}.

Therefore reconstruction networks can not be used to explain the activation of capsules, because they are simply trained to minimize the distance between the current input and the output independent on what a capsule really represents.




\section{Activations of capsules for generated input features}
In the paper we reported images that activate single output units.
In \cref{tbl:activations_of_gen} we show the activation's $||v_j||$ of each capsule that is produced for each capsule:
\begin{table}[H]
   \begin{center}
   \begin{tabular}{c|cccc}
   \toprule
    Capsule & SDA-$5$k & SDA-$15$k & RBA-$15$k \\
   \midrule
    0 & 0.84 & 0.85 & 0.99 \\
    1 & 0.80 & 0.89 & 0.99 \\
    2 & 0.71 & 0.81 & 0.99 \\
    3 & 0.85 & 0.89 & 0.99 \\
    4 & 0.65 & 0.70 & 0.99 \\
    5 & 0.89 & 0.93 & 0.99 \\
    6 & 0.40 & 0.62 & 0.99 \\
    7 & 0.40 & 0.59 & 0.99 \\
    8 & 0.88 & 0.92 & 0.99 \\
    9 & 0.90 & 0.92 & 0.99 \\
   \bottomrule
   \end{tabular}
   \end{center}
   \caption{Activations of the output capsules for the images that are generated in the paper.}
   \label{tbl:activations_of_gen}
\end{table}
As we can see after $5$k steps (SDA) for capsule $6$ and $7$ generated input features produce a relative low activation of the output indicating that the network is still not confident about the features that it learned. We can see that after $15$k steps (SDA) this confidence increased and it learned features for class $6$. Also the confidence for class $7$ increased although we have seen that not really comprehensible features where found indicating that the class represents \emph{none-of-the-above} rather than sneakers. We also want to outline that for RBA the activation is very large, although only single pixels where activated. This indicates that model that rely on non-robust useful features are very confident about their predictions. On the other hand for SDA features are much more comprehensible and the model is much more conservative for making predictions.

\section{Convolutional filter of $\gamma$-capusle networks}
\begin{figure}[H]
   \begin{center}
      \includegraphics[width=0.45\textwidth]{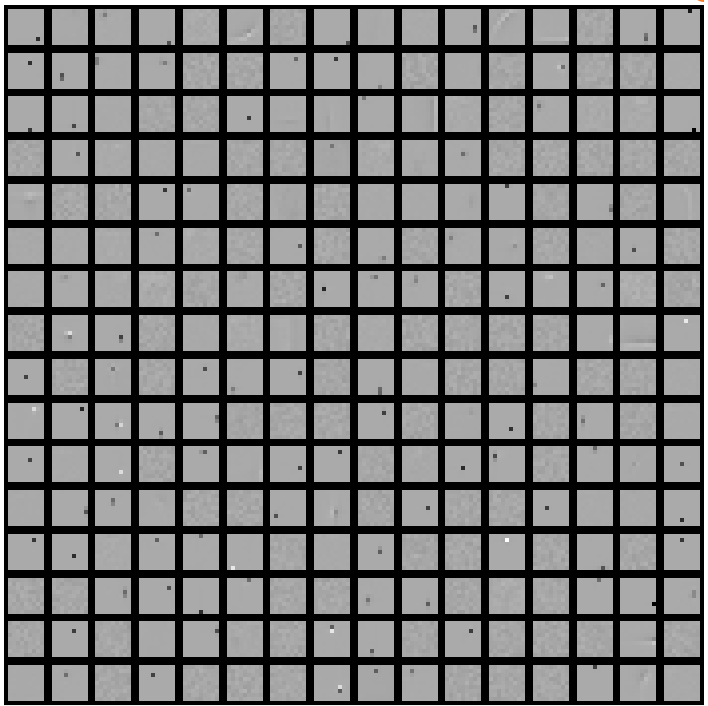}
   \end{center}
      \caption{All $256$ convolutional filters learned by $\gamma$-capsule networks.}
   \label{fig:conv_filter}
\end{figure}

\citet{resistant-to-adversraial-attacks} reports that convolutional filter of robust models have significantly more concentrated weights. For $\gamma$-capsule networks we have seen that similar convolutional filters are learned \cite{resistant-to-adversraial-attacks} with robust training. All $256$ filters are shown in \cref{fig:conv_filter}. 

We conclude that the additional robustness under strong attack cannot be due to the convolutional filters. It is also not a property of capsule networks in general as shown by \cite{capsule-adversarial-attacks} and therefore we claim that it is a property of $\gamma$-capsule layers of the $\gamma$-capsule network.



\end{document}